# Dental Panoramic Radiograph Analysis Using YOLO26: From Tooth Detection to Disease Diagnosis

**Khawaja Azfar Asif [a], Rafaqat Alam Khan [a]**

[a] Department of Software Engineering, Lahore Garrison University, DHA Phase 6 Sector C, Lahore, Pakistan

sp20-bsse-023@lgu.edu.pk, rafaqatalam@lgu.edu.pk

**Abstract:**
This study aimed to develop and evaluate an automated method for teeth enumeration and dental disease segmentation on panoramic radiographs using YOLO26 models. Addressing the limitations of manual teeth counting, which is tedious, error-prone, and impractical in busy clinical settings, we investigated whether YOLO26-based models could outperform the YOLOv8x baseline in detection accuracy and segmentation performance. The DENTEX dataset was preprocessed using Roboflow for format conversion and augmentation. A total of 1,082 images were used for teeth counting and 1,040 images for disease segmentation across four pathology categories. Five YOLO26-seg models (n-seg, s-seg, m-seg, l-seg, x-seg) were trained on Google Colab with transfer learning at 800×800 resolution. Results showed that YOLO26-m-seg achieved superior performance for teeth enumeration with precision 0.976, recall 0.970, box mAP50 0.976, and mask mAP50 0.970, outperforming YOLOv8x by 4.9% in precision and 3.3% in mAP50. For disease detection, YOLO26-l-seg reached box mAP50 of 0.591 and mask mAP50 of 0.547, with the highest precision for impaction (0.943). YOLO26 models, particularly the m-seg variant, offer a clinically viable tool for rapid and accurate panoramic radiograph analysis, reducing manual workload and supporting efficient treatment planning.

## 1. Introduction:

Oral diseases currently affect around 3.5 billion individuals, making them some of the most common non-communicable diseases in the world [1]. Panoramic radiography, or orthopantomography, is the basic imaging technique used in dentistry [2]. A single panoramic image includes both jaws, the temporomandibular joint areas, and adjacent anatomical regions, thus providing an efficient diagnostic examination in a matter of seconds and reducing the radiation dose [3]. While panoramic images prove their effectiveness in clinical practice, reading them is considered a challenging mental task. Dentists must recognise and label all 32 teeth according to the standard FDI World Dental Federation nomenclature [4] while diagnosing periodontal diseases, identifying potential pathological conditions, and assessing bone level. Moreover, these actions need to be performed in a single image where various structures might overlap and have variable image quality, which leads to difficulties in recognizing some conditions, especially at their initial stages. Research indicates that there are considerable differences between observers in interpreting dental radiographs, mainly concerning detecting caries and periapical lesions [5], [6]. In high-load clinical settings like dental teaching hospitals and population dental screening programs, such differences become apparent and affect patients negatively.

The development of deep learning technologies presents new opportunities for analyzing dental images automatically. Recent research shows the outstanding performance of CNNs and Transformer models for solving detection, segmentation, and classification tasks in various medical imaging modalities [7],[8]. In dental AI, the ability to concurrently detect all teeth in a dental image, label them according to FDI notation, and find any signs of pathology is especially useful from the clinical point of view. However, no work to date addresses these issues simultaneously.

Currently, the most efficient architecture for real-time medical imaging detection is the family of YOLO detectors [10]. Introduced in January 2026 by Ultralytics, YOLO26 is one of the best versions due to its numerous advantages: NMS-free inference allows removing latency associated with post-processing, ProgLoss facilitates the process of converging to balanced results on imbalanced datasets, STAL significantly improves detection of small objects, and the MuSGD optimiser enables stable training[10]. The presented features are especially important for dental radiographs because different teeth can vary significantly in size, and disease lesions may be small, subtle, and overlapped. The following are the major contributions made by this paper: (1) the first implementation of YOLO26 in the field of dental panoramic radiograph analysis; (2) a detailed assessment of all five YOLO26-seg model variants used in both teeth enumeration and multi-class disease segmentation tasks; (3) outstanding results obtained on the DENTEX dataset, outperforming the YOLOv8x baseline by 4.9% in terms of precision; (4) a pioneering pipeline for solving both teeth enumeration and multi-class disease segmentation problems on DENTEX dataset in a unified way, including the automated preprocessing workflow in Roboflow; and (5) unique class-level analysis revealing visual distinctiveness as the key determinant of detection performance, regardless of annotation quantity.

## 2. Related Work:

Over the last few years, considerable progress has been made in automating the processing of dental radiographs, moving away from conventional computer vision techniques towards the use of more powerful deep learning techniques. The initial research relied heavily on pre-existing hand-crafted features, thresholding, and morphological segmentation techniques that lacked generalizability owing to the variability of panoramic radiographs among other factors [11]. It is safe to say that the advent of convolutional neural networks revolutionized this field of research and made it possible to learn useful features from scratch starting from raw radiographic images. One of the most notable contributions is by Jader et al. [12], who implemented a Mask R-CNN framework with transfer learning from pre-trained models based on the MS-COCO dataset to perform panoramic dental radiograph segmentation [29]. This study proved that deep instance segmentation can be used to segment individual tooth boundaries, achieving an average precision of 0.78 per tooth class. Still, it considered teeth segmentation only and had no regard for FDI-based tooth numbering and pathological cases.

For tooth recognition, Mahdi et al. [13] implemented the Faster R-CNN algorithm with two pre-trained ResNet backbones (ResNet-50 and ResNet-101) and incorporated K-fold cross-validation. Although they demonstrated good results in recognizing individual teeth, they did not include any disease detection in their algorithm. To investigate the problem of missing teeth to facilitate implant planning, Park et al. [14] employed Mask R-CNN and achieved an AUC of 0.94. Still, this solution was specific to the missing teeth case, not considering any other types of diseases. On the topic of dental disease detection, Lee et al. [15] employed convolutional neural networks to detect caries in bitewing radiographs and achieved promising results for individual carious lesions. Yet, they pointed out the significant difficulty of generalising the solution for whole panoramic radiographs with multiple overlapping tooth structures. Cantu et al. [16] developed a deep learning-based approach for proximal caries detection and explicitly stated severe class imbalance issues in their dataset. This limitation is crucial and became a driving force for dataset generation in the present study. Finally, the development of the DENTEX dataset [31] introduced the first annotated panoramic dental radiograph dataset for simultaneous teeth enumeration and disease detection. Specifically, most relevant to the current research, Mendes et al. [18] implemented the YOLOv8x model [25,26,27] to perform automated tooth detection with subsequent FDI numbering on the DENTEX dataset and achieved a precision of 0.9283 and an mAP50 of 0.9450. Importantly, they noted disease detection as one of the potential areas for future improvements. As far as we know, there have been no studies that utilized YOLO26 for dental radiograph processing, let alone those that simultaneously enumerated individual teeth and segmented dental diseases at once on the DENTEX dataset. Table 1 summarizes the related works in the context of the present study.

Table 1. Summary of related work in automated dental radiograph analysis

| Study | Method | Dataset | Task | Metric | Limitation |
|---|---|---|---|---|---|
| Jader et al. [12] 2018 | Mask R-CNN | Custom panoramic | Tooth segmentation | AP: 0.78 | No numbering, no disease |
| Mahdi et al. [13] 2020 | Faster R-CNN ResNet | Custom panoramic | Tooth recognition | Acc: 0.91 | No disease detection |
| Park et al. [14] 2022 | CNN-based | Custom panoramic | Missing tooth detection | AUC: 0.94 | Single pathology only |
| Lee et al. [15] 2018 | CNN | Bitewing X-ray | Caries detection | AUC: 0.94 | Not panoramic view |
| Cantu et al. [16] 2020 | Deep CNN | Bitewing X-ray | Proximal caries | AUC: 0.89 | Class imbalance issue |
| Mendes et al. [18] 2024 | YOLOv8x | DENTEX | Tooth detection + FDI | mAP50: 0.945 | No disease detection |
| Ours (2026) | YOLO26-seg | DENTEX | Teeth + Disease | mAP50: 0.976 | First YOLO26 dental study |

## 3. Materials and Methods:

### 3.1. Proposed Methodology:

Figure 1 demonstrates the entire pipeline of the proposed system. As a first step, the DENTEX dataset is acquired from the public source Zenodo, provided in COCO JSON format. Notably, COCO is incompatible with YOLO frameworks used to develop object detection models. Thus, the next step involves the pre-processing and conversion of the COCO-format images to the YOLO26 format via the Roboflow platform. Images are automatically processed through the following steps: COCO to YOLO26 format conversion, automatic EXIF stripping to correct the images' orientation, and automatic resizing of the images to the standardized resolution of 2048×1010. Data augmentation is not applied, as original clinical information is maintained.

Next, the preprocessed dataset is split into two different tasks: teeth enumeration and disease segmentation. Teeth enumeration comprises 1,082 images grouped into four classes (canine, incisors, molars, and premolars), with data split into 890 for training, 70 for validation, and 122 for testing. Disease segmentation includes 1,040 images with four classes (caries, deep caries, impacted teeth, and periapical lesions), and the split between training (845), validation (40), and testing (155) images.

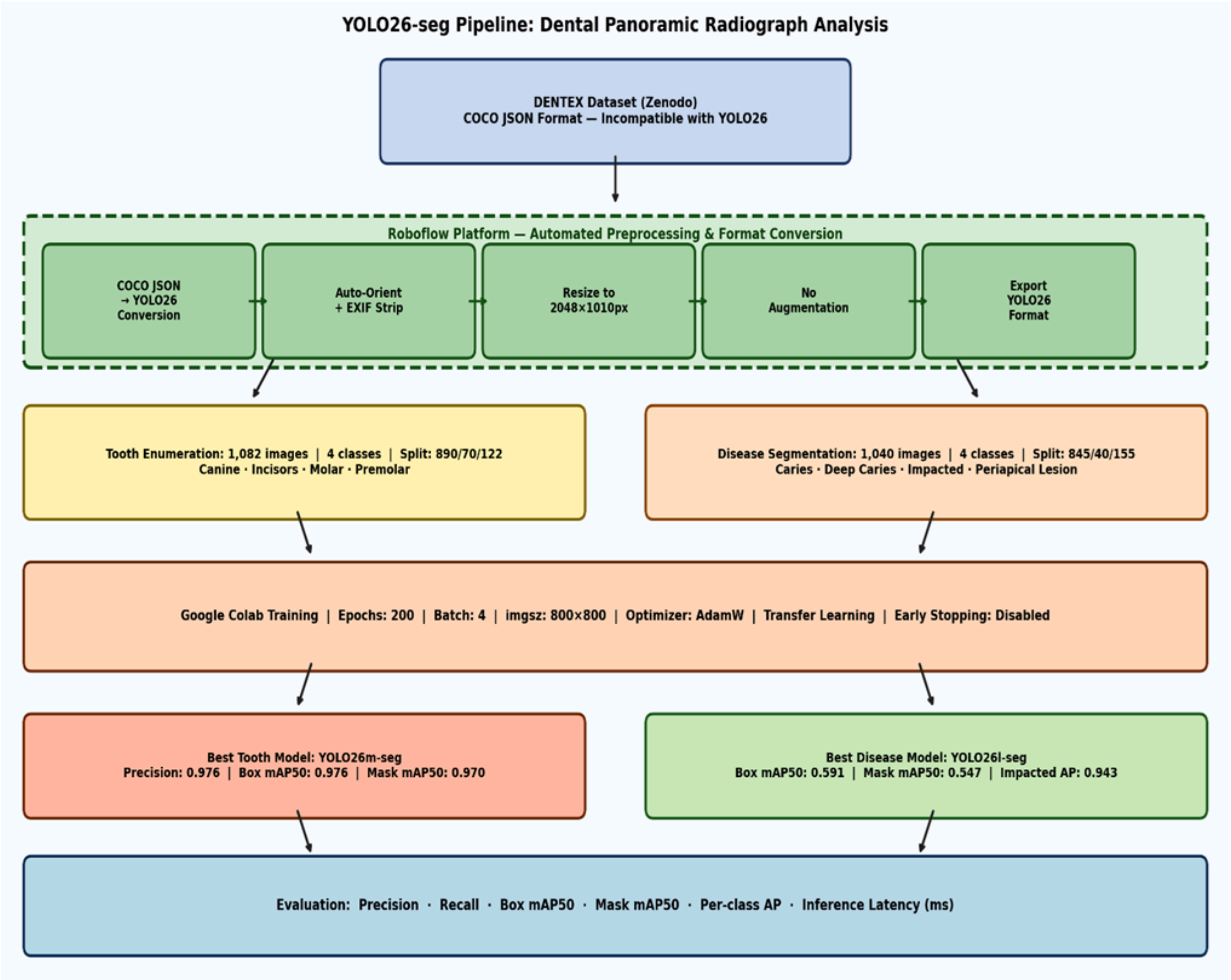

Figure 1. Proposed YOLO26-seg pipeline for automated dental panoramic radiograph analysis, showing dataset acquisition, Roboflow preprocessing, model training configuration, and dual-task evaluation outputs.

### 3.2. Dataset

The DENTEX dataset [31] is currently the largest publicly accessible database for panoramic dental X-ray image analysis that contains hierarchical annotations with three different levels, namely, quadrant detection, teeth detection with FDI notations, and disease detection. The initial form of the DENTEX database has been provided in COCO JSON format [28] for Detectron2 and does not fit into YOLO-based training processes. That is why the DENTEX database was imported to the Roboflow platform [19], and the process of conversion from COCO JSON to YOLO26 format annotations was done successfully.

As part of the first task involving the detection of teeth on panoramic radiographs, a dataset containing 1,082 panoramic radiographs with 30,732 annotations across four FDI teeth classes (Canine, Incisors, Molar, Premolar) was prepared. The dataset has been split into training, validation, and testing subsets: 890 images (82.3%), 70 images (6.5%), and 122 images (11.3%), respectively. The distribution of teeth annotations reflects human dental anatomy. Molars are the most frequently represented class (33.3%, up to 12 per patient), while canines are the least represented (13.9%, 4 per patient). This clinically realistic distribution supports the suitability of the dataset for training models with strong generalization capability.

The disease segmentation dataset consists of 1,040 images with 5,428 annotations distributed across four pathology categories: Caries, Deep Caries, Impacted Teeth, and Periapical Lesion. The dataset has been split into training, validation, and testing subsets: 845 images (81.3%), 40 images (3.8%), and 155 images (14.9%), respectively. The disease dataset shows a pronounced class imbalance, where Caries accounts for 62.8% of all annotations, while Periapical Lesion represents only 4.3%. This corresponds to a 14.7× imbalance ratio, consistent with its lower clinical prevalence in real-world settings.

We used the FDI two-digit notation system for tooth identification, in accordance with international clinical standards [20]. In this system, the first digit indicates the quadrant (1–4 for permanent dentition and 5–8 for primary dentition), while the second digit specifies the tooth position within the quadrant (1–8, starting from the midline). In this study, the four tooth-type classes are as follows: Incisors, Canines, Premolars, and Molars.

### 3.3. YOLO26 Architecture

YOLO26, released by Ultralytics in January 2026 [10], builds on the YOLOv8 backbone with four key architectural improvements that are particularly relevant to dental radiograph analysis.

- **NMS-Free Inference:** Eliminates the non-maximum suppression stage by adopting an end-to-end detection head, reducing inference latency and preventing suppression of closely spaced detections. This is especially important in panoramic radiographs, where adjacent teeth are densely packed and may otherwise be incorrectly filtered out.
- **Progressive Loss Balancing (ProgLoss):** Dynamically reweights classification, bounding box regression, and segmentation losses during training. This is particularly beneficial for disease segmentation, where a strong class imbalance (14.7× ratio) could otherwise bias learning toward the dominant Caries class.
- **Small-Target-Aware Label Assignment (STAL):** Enhances label assignment for small and overlapping objects. This directly improves detection of fine structures such as early caries.

- **MuSGD Optimizer:** A momentum-based stochastic gradient descent variant that ensures stable convergence under imbalanced and heterogeneous data distributions, which is especially important for learning from underrepresented disease classes.

Five model variants were evaluated in this study: YOLO26-n-seg (nano: 2.69M parameters, 9.0 GFLOPs), YOLO26-s-seg (small: 10.37M parameters, 34.1 GFLOPs), YOLO26-m-seg (medium:23.51M parameters, 121.2 GFLOPs), YOLO26-l-seg (large: 27.91M parameters, 139.4 GFLOPs), and YOLO26-x-seg (extra-large: 70.49M parameters, 336.7 GFLOPs).

### 3.4. Experimental Setup

All experiments were performed on Google Colab, utilizing NVIDIA GPU hardware. Training artifacts were stored in Google Drive. Both models were initialized with official YOLO26 pretrained weights, allowing transfer learning from large-scale object detection pretraining to dental images [9].
In both cases, training was done with the same hyperparameters, which included 200 epochs, a batch size equal to 4, and an input resolution of 800×800 pixels. Despite the fact that the original Roboflow dataset had a resolution of 2048×1010, images were downscaled to 800×800 due to GPU memory limitations on Colab. As for the optimizer, AdamW with the YOLO26 default scheduling scheme was used. In addition, early stopping was not applied to ensure the full 200-epoch training process is completed. In case of any session interruption, training could be resumed using the resume=True flag.

While this study uses distinct YOLO26 models for detection and segmentation, the baseline uses YOLOv8 only for detection, with no segmentation or disease detection attempted. Furthermore, the proposed model is based on a larger dataset and makes use of Roboflow automation to convert the images into a proper format. More importantly, training schedules and resolutions are better optimized in the proposed experiment. Finally, multi-class disease detection was implemented as opposed to single-class detection used in prior studies.

Precision is the ratio of correct detections to all predictions made by the network. On the other hand, Recall refers to the number of true positives relative to the total number of true objects in an image. Box mAP50 stands for the mean average precision measured at IoU = 0.50 for the bounding box predictions. Likewise, Mask mAP50 refers to the same metric for mask predictions. Inference latency (milliseconds per image) estimates the efficiency of the inference time. It is critical to consider whether the model is suitable for practical purposes. Besides that, the per-class Average Precision (AP) has been estimated based on Precision–Recall curves [21].

## 4. Results and Discussion

### 4.1. Dental Enumeration, YOLO26m-seg

Table 2 describes the results for all five versions of YOLO26-seg used to detect teeth in panoramic radiography images. YOLO26-m-seg achieved the best results with maximum precision (0.976), recall (0.970), and mAP50 (0.976 for box, 0.970 for mask). Lightweight models like YOLO26-n-seg performed similarly with lower computational cost, while more complex models like YOLO26-x-seg showed no improvement despite increased complexity.

Table 2. YOLO26-seg variant performance for dental enumeration (best results highlighted)

| Model | Params(M) | GFLOPs | Precision | Recall | Box mAP50 | Mask mAP50 | Latency(ms) |
|---|---|---|---|---|---|---|---|
| YOLO26-n-seg | 2.69 | 9.0 | 0.966 | 0.958 | 0.966 | 0.958 | 22.8 |
| YOLO26-s-seg | 10.37 | 34.1 | 0.961 | 0.957 | 0.961 | 0.957 | 27.1 |
| **YOLO26-m-seg** | **23.51** | **121.2** | **0.976** | **0.970** | **0.976** | **0.970** | **29.4** |
| YOLO26-l-seg | 27.91 | 139.4 | 0.971 | 0.960 | 0.971 | 0.960 | 25.2 |
| YOLO26-x-seg | 70.49 | 336.7 | 0.956 | 0.945 | 0.956 | 0.945 | 25.4 |

YOLO26-m-seg outperforms all models, scoring 0.976, 0.970, 0.976, and 0.970 for precision, recall, box mAP50, and mask mAP50, with 29.4 ms inference time, making it suitable for clinical real-time teeth detection. YOLO26-n-seg achieves good precision of 0.966 in 22.8 ms, useful for mobile applications. Conversely, YOLO26-x-seg, with a higher number of parameters, showed poor performance, indicating that simply enlarging model capacity does not improve results for teeth enumeration.

Figure 2 shows the confusion matrix for teeth classification. All diagonal values are high for each class: 0.99 for incisors, 0.97 for molars, 0.96 for canines, and 0.96 for premolars, respectively. Classification performance is therefore high for all types of teeth with negligible confusion. False positives occur mainly when some background pixels are erroneously classified as teeth. This is expected given the density of the jaw structures found in panoramic radiographs.

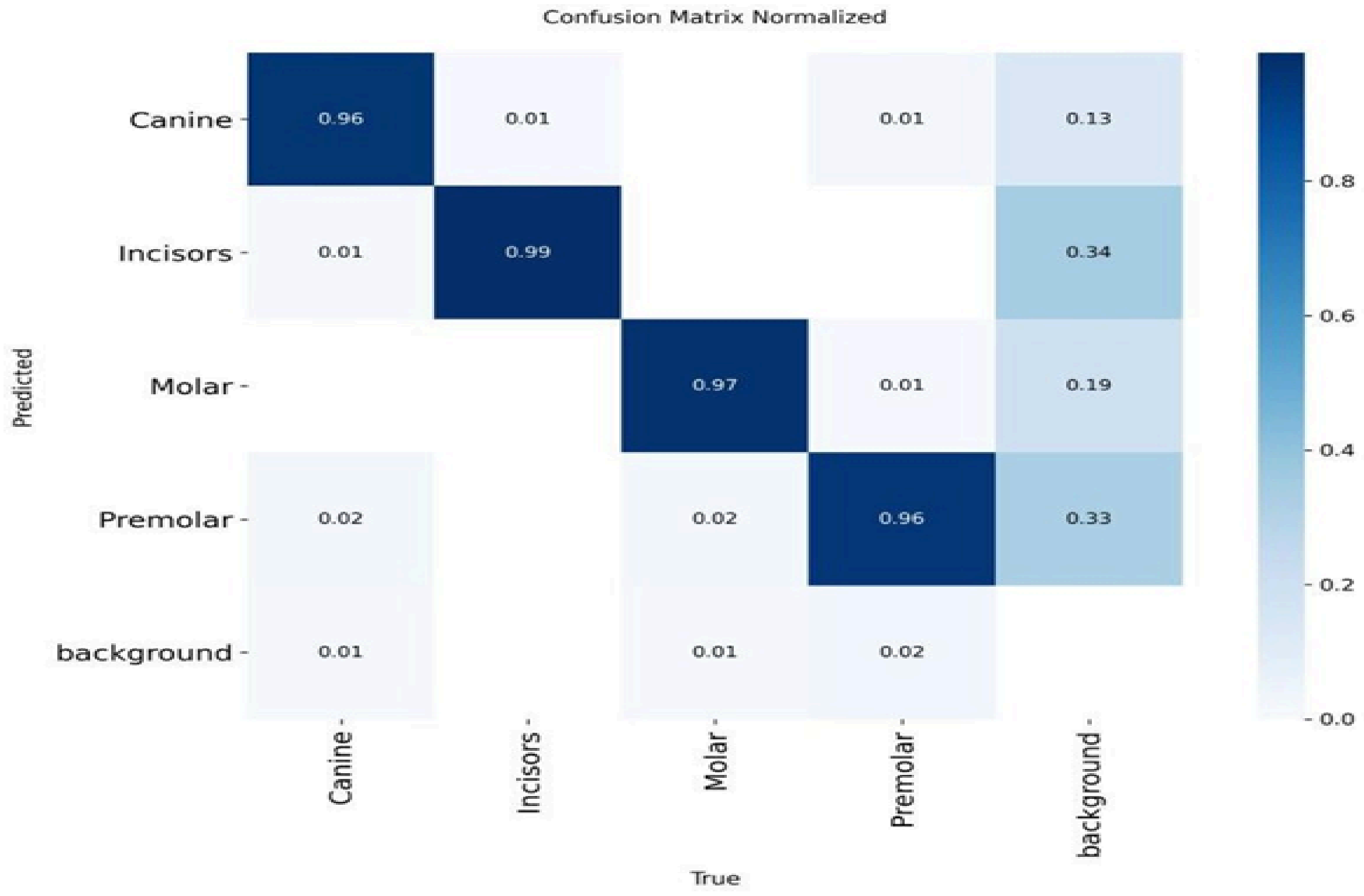

Figure 2. Confusion matrix for YOLO26m-seg,  Dental Enumeration Task.

Figure 3 depicts qualitative inference results on test images not seen by the model during training. It precisely detects and classifies all present teeth, including molars, premolars, canines, and incisors, with confidence scores between 0.80 and 0.99.

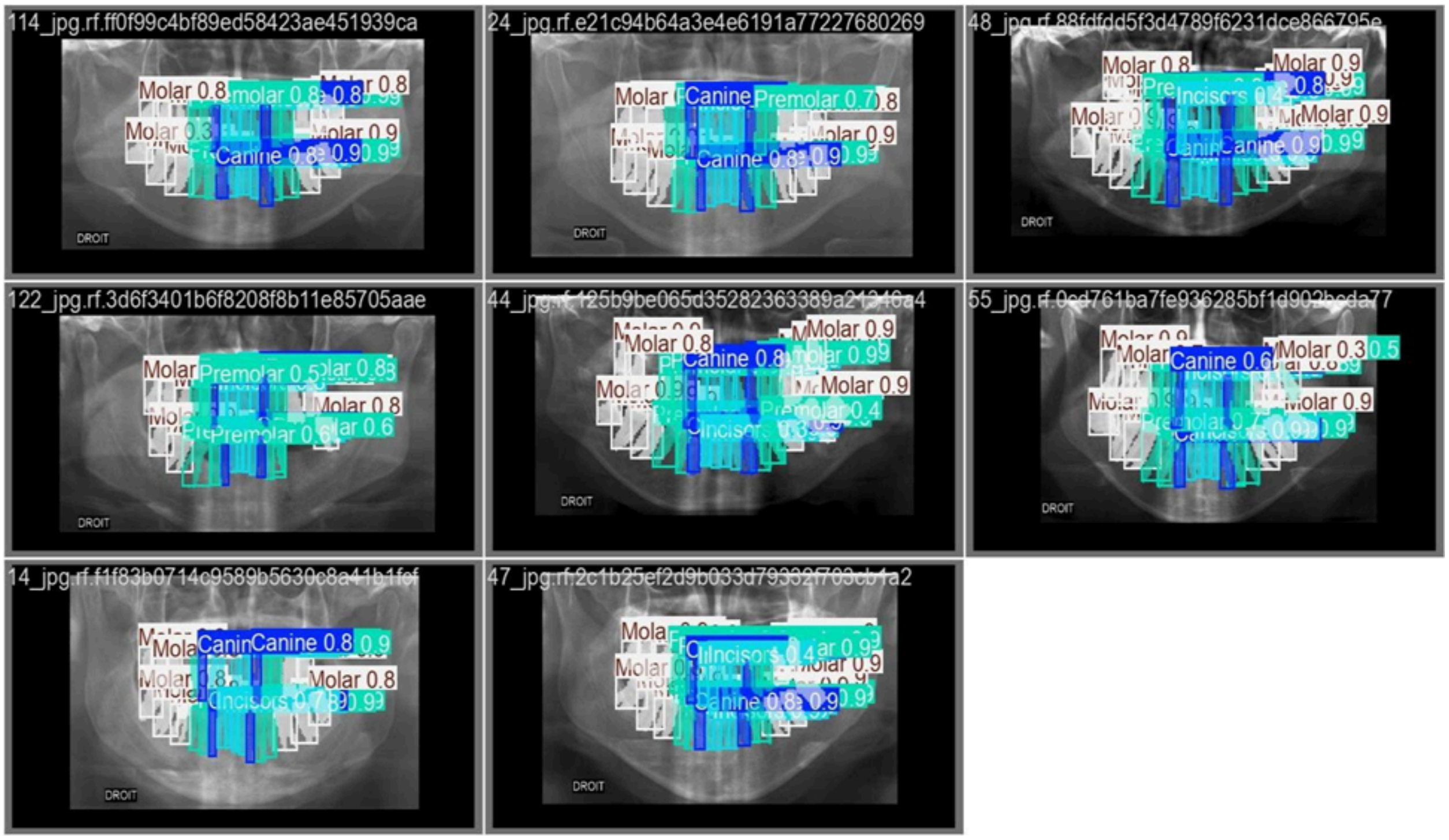

Figure 3. YOLO26m-seg qualitative teeth detection predictions on held-out test images.

### 4.2. Dental Disease Segmentation, YOLO26-l-seg

Table 3 describes the results for all five versions of YOLO26-seg used to detect disease in panoramic radiography images. YOLO26-l-seg achieved the highest box and mask mAP50 (0.591 and 0.547), demonstrating the best balance between detection and segmentation accuracy. While YOLO26-s-seg and YOLO26-x-seg show good precision, their overall performance is inferior to YOLO26-l-seg. YOLO26-n-seg is faster and more efficient but with considerably lower accuracy due to its inability to capture complex dental disease patterns. Therefore, larger network variants better recognize complex disease patterns and provide superior segmentation, making YOLO26-l-seg the optimal choice.

Table 3. YOLO26-seg variant performance for dental disease segmentation (best results highlighted)

| Model | Params(M) | GFLOPs | Precision | Recall | Box mAP50 | Mask mAP50 | Latency(ms) |
|---|---|---|---|---|---|---|---|
| YOLO26-n-seg | 2.69 | 9.0 | 0.524 | 0.477 | 0.480 | 0.441 | 16.5 |
| YOLO26-s-seg | 10.37 | 34.1 | 0.683 | 0.551 | 0.547 | 0.489 | 21.7 |
| YOLO26-m-seg | 23.51 | 121.2 | 0.521 | 0.595 | 0.562 | 0.484 | 27.3 |
| **YOLO26-l-seg** | **27.91** | **139.4** | **0.603** | **0.611** | **0.591** | **0.547** | **43.9** |
| YOLO26-x-seg | 70.49 | 336.7 | 0.654 | 0.527 | 0.582 | 0.519 | 40.2 |

YOLO26-l-seg achieved the highest disease segmentation performance (box mAP50 = 0.591, mask mAP50 = 0.547), indicating that larger models are better suited for disease detection than teeth detection. This is logical given the complexity of disease recognition.

Class imbalance significantly affected performance. Caries, the largest class (62.8%, 3,409 examples), had low AP (0.463), while Impacted Teeth (17.1%, 927 examples) achieved high AP (0.943) due to distinctive visual characteristics. Periapical Lesions performed worst (AP = 0.384) with only 4.3% prevalence, likely due to their faint and indistinct radiographic appearance.

Figure 4 shows the Confusion Matrix for disease classification across four categories. Impacted lesions demonstrated the highest accuracy (0.92 diagonal coefficient) due to their clear visibility on panoramic radiographs. Conversely, Caries and Periapical Lesions showed the lowest accuracy (0.42 and 0.10), primarily because they were confused with Background (0.58 and 0.80, respectively). These conditions' low visibility and underrepresentation in the dataset explain the detection failures.

Deep Caries achieved 0.78 correct classification but was frequently confused with Caries (0.14) and Impacted (0.06) due to structural similarity. Overall, lesion visibility and structure are more influential than dataset frequency.

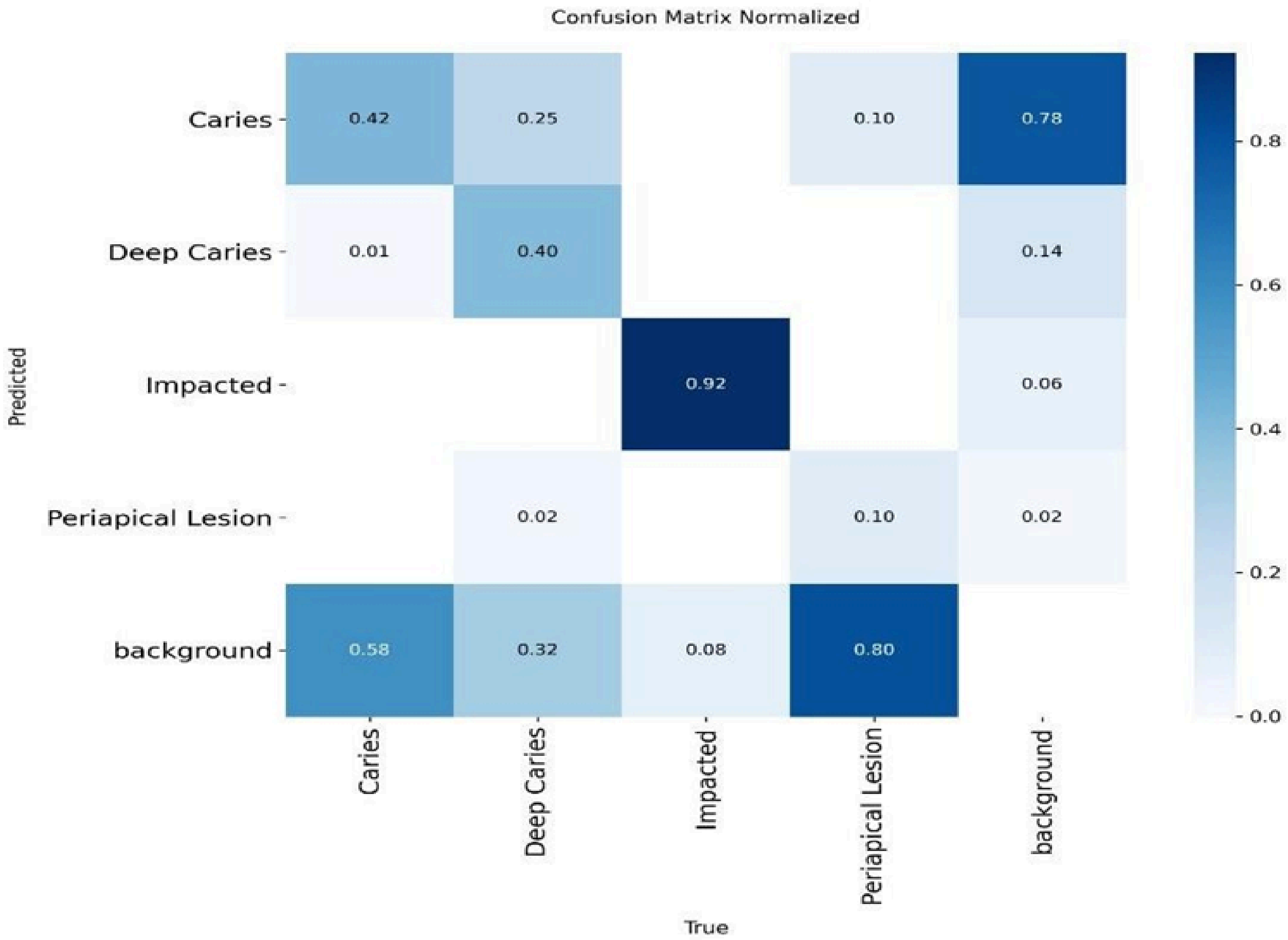

Figure 4. Confusion matrix for YOLO26-l-seg, Disease Segmentation Task.

Figure 5 demonstrates qualitative outcomes of disease segmentation obtained by the proposed model. Thus, the results demonstrate the ability of the model to detect and localize various dental pathologies through segmentation masks for different clinical cases. Furthermore, the model proves to be able to detect several pathologies at once in cases of their co-occurrence, such as tooth decay along with impaction of teeth. Nevertheless, some undetected pathologies in areas of low contrast indicate that further improvement is necessary to overcome the current limitation.

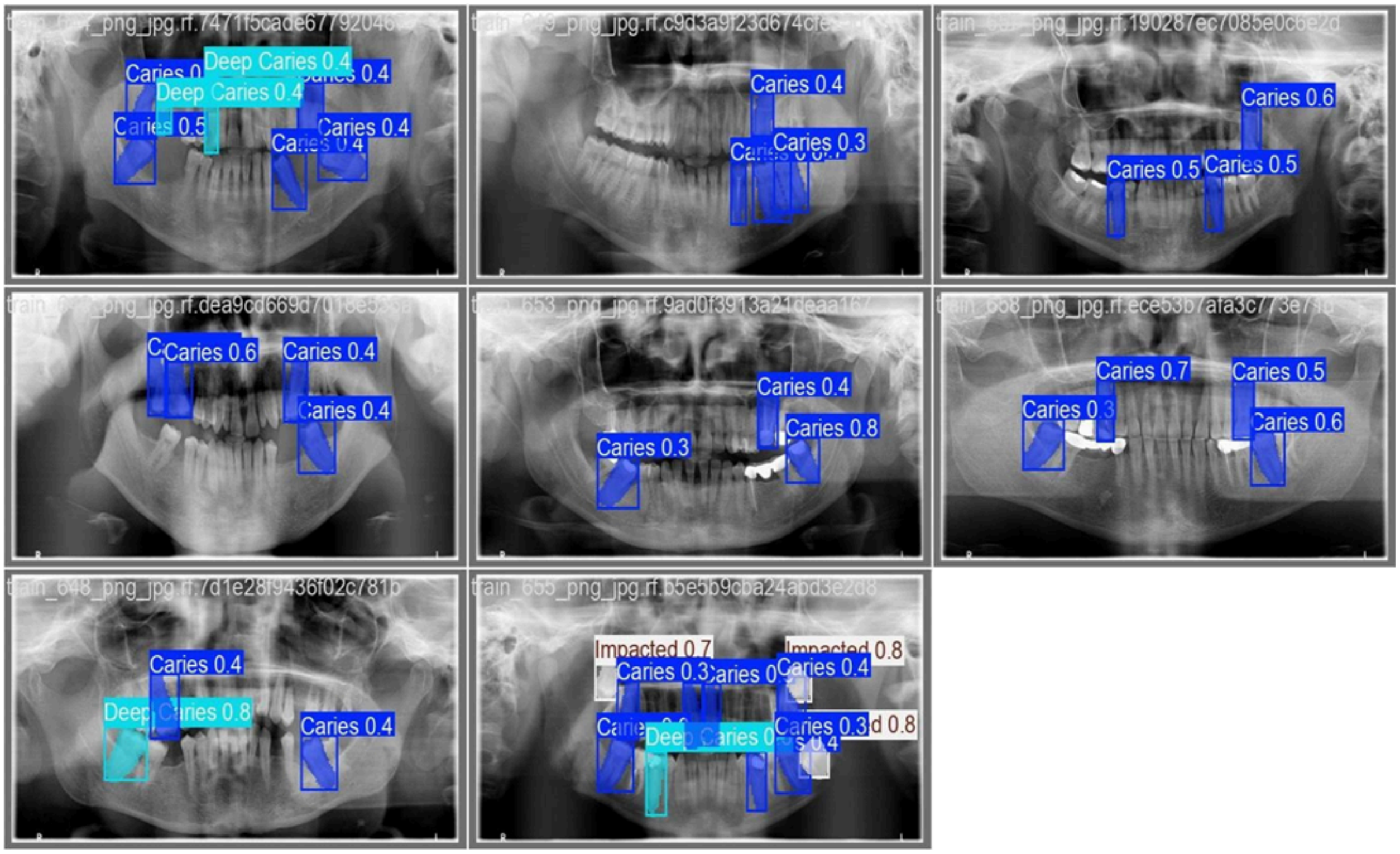

Figure 5. YOLO26-l-seg qualitative disease segmentation predictions on held-out test images. Segmentation masks with disease-type labels and confidence scores demonstrate successful multi-class pathology detection, including co-occurring conditions.

## 4.3. Comparison with Studies Using DENTEX Dataset

Tables 4 and 5 compare our proposed YOLO26-based models with previous work on the DENTEX dataset. For teeth detection in dental radiographs, these tables include the original DENTEX paper by Hamamci et al. [17], which used Faster R-CNN to detect and number teeth according to FDI notation, and a recent study that applies YOLOv8x by Mendes et al. [18]. All the above methods focused only on teeth detection and numbering without performing disease segmentation.

**Teeth Enumeration Task:**

Regarding the teeth enumeration task, the proposed YOLO26-m-seg outperformed all previously reported methods. Specifically, compared to Mendes et al. [18], where they reached 0.928 for the precision score and 0.945 for the mAP50 metric when using YOLOv8x, our model achieved a precision score of 0.976 and 0.976 for mAP50. The respective improvements compared to Mendes et al. [18] amount to 4.9% and 3.3%. As can be seen in Figure 6, YOLO26-m-seg demonstrates superior performance compared to YOLOv8x. Furthermore, YOLO26-m-seg offers mask-level segmentation (mAP50 = 0.970), which was not possible in previous models. Table 4 compares different studies.

Table 4. Comparison with published studies of teeth detection using the DENTEX dataset

| Study | Model | Task | Precision | mAP50 |
|---|---|---|---|---|
| Hamamci et al. [17] 2023 | Faster R-CNN | Tooth detection | 0.891 | 0.871 |
| Mendes et al. [18] 2024 | YOLOv8x | Tooth detection | 0.9283 | 0.9450 |
| **Our** | **YOLO26-m-seg** | **Teeth detection** | **0.976** | **0.976** |

**Disease Segmentation Task:**

Regarding disease segmentation, no previous study using the DENTEX dataset reports results for this task. Thus, YOLO26-l-seg became the first model to perform dental pathology segmentation on this benchmark. YOLO26-l-seg reaches a precision score of 0.603 and box mAP50 of 0.591 for four types of pathologies: caries, deep caries, impacted, and periapical lesions. Although results are modest, we believe that they will serve as a baseline for future investigations in this direction. Table 5 compares different studies.

Table 5. Comparison with published studies of disease segmentation using the DENTEX dataset

| Study | Model | Task | Precision | mAP50 |
|---|---|---|---|---|
| Hamamci et al. [17] 2023 | Faster R-CNN | Tooth detection | 0.891 | 0.871 |
| Mendes et al. [18] 2024 | YOLOv8x | Tooth detection | 0.9283 | 0.9450 |
| **Our** | **YOLO26-l-seg** | **Disease Segmentation** | **0.603** | **0.591** |

Apart from detecting and counting teeth, YOLO26-l-seg offers disease segmentation capabilities not studied in the context of DENTEX before. It provides the first results in this field, achieving baseline values of box mAP50 and mask mAP50 (0.591 and 0.547, respectively) on four clinically relevant disease categories. These additional data may be beneficial for treatment planning, dental prosthetics, or orthodontic diagnostics.

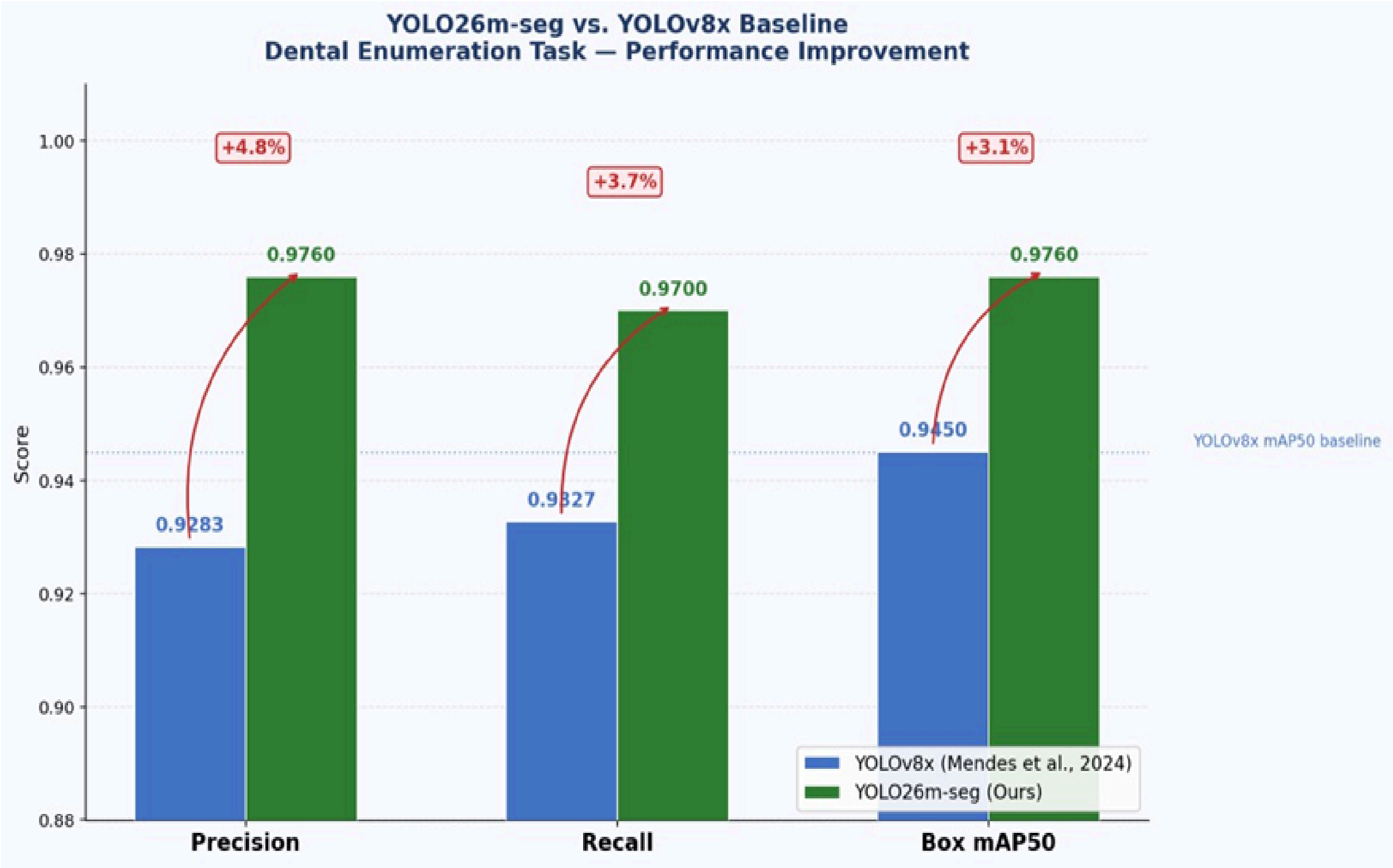

Figure 6. Performance comparison of YOLO26-m-seg vs. YOLOv8x baseline (Mendes et al. [18]) on dental enumeration: Precision +4.9%, Recall +4.0%, Box mAP50 +3.3%, with additional Mask mAP50 0.970 introduced.

### 4.4. Limitations

There are some limitations that should be considered in order to put the findings into the right perspective. Firstly, all the experiments are done in one training session, without any statistical analysis of several independent training sessions due to the restrictions of Google Colab [30] and limited GPU availability. Secondly, the problem with the data is that there is a high degree of imbalance between classes, since the Periapical Lesion makes up only 4.3% of the total number of disease images. Thirdly, because of the limited GPU resources, training can be done only at 800×800 resolution instead of the original 2048×1010 provided by Roboflow. Finally, the experiments have been done only with the data from the DENTEX benchmark, and further investigations are needed to see how it performs in radiographs obtained from other manufacturers.

## 5. Conclusion and Future Work

The study described here reported the first use of YOLO26 on the problem of automated dental panoramic radiograph interpretation, setting new state-of-the-art performance on the benchmark dataset DENTEX in terms of teeth detection and dental disease segmentation. Specifically, five versions of YOLO26-seg have been rigorously benchmarked for two distinct tasks based on datasets containing 1,082 images for teeth enumeration and 1,040 images for disease segmentation. These datasets have been produced using the new Roboflow pipeline, as compared to manual approaches used in previous work. For teeth enumeration, YOLO26-m-seg reached a precision of 0.976 and a box mAP50 of 0.976, improving the baselines set by YOLOv8x by 4.9% and 3.3%, respectively, and further adding segmentation performance (mask mAP50 of 0.970), which was not previously addressed in any DENTEX study. For the problem of disease segmentation, considering four clinically relevant pathology types (Caries, Deep Caries, impacted teeth, and Periapical Lesion), YOLO26-l-seg was able to reach box mAP50 of 0.591 and mask mAP50 of 0.547, representing the first quantifiable baseline for such a task on the DENTEX benchmark. The main conclusion drawn from the current work is that visual uniqueness predicts performance better than annotation number. Thus, impacted teeth reached an AP of 0.943 despite only 927 annotated images since they have a very distinctive appearance. On the other hand, Caries only scored an AP of 0.463 despite 3,409 instances due to low visual distinguishability and confusion with Deep Caries. This has important implications in the context of dataset and model creation for dental AI research, investing resources in annotation quality may produce better performance improvements than annotation number, especially when dealing with visually challenging diseases.

In clinical practice, the system described here has great potential in supporting dentists by helping to reduce radiograph evaluation time, avoiding inconsistencies in the process of teeth numbering, and allowing automatic detection of pathology requiring further investigation by clinicians. All variants can achieve sub-45 ms inference latency, thus guaranteeing the ability to perform online deployment. Future work will follow four main lines: (1) addressing class imbalance by collecting additional data and using targeted augmentations for rare pathologies such as Periapical Lesion; (2) training models at the full 2048 × 1010 resolution on more performant hardware to benefit from fine-grained details in small lesions detection; (3) external testing on multi-center datasets featuring different radiography devices from distinct manufactures; and (4) development of a decision support tool integrated into clinical practice based on prospective evaluation of dental professionals.

**Acknowledgements**
The authors acknowledge Google Colab for providing computational resources, Roboflow for dataset preprocessing and annotation conversion tools, and the DENTEX dataset creators for making the benchmark publicly available to the research community.

**Correspondence Author:** Khawaja Azfar Asif

**Conflict of Interest**
Khawaja Azfar Asif and Rafaqat Alam Khan declare no conflict of interest.

**Ethical Approval**
This article does not contain any studies with human participants or animals performed by any of the authors.

**Informed Consent**
Informed consent was not required for this study.